%% file: conference_101719.tex
\documentclass[conference]{IEEEtran}
\IEEEoverridecommandlockouts
\usepackage{cite}
\usepackage{amsmath,amssymb,amsfonts}
\usepackage{algorithmic}
\usepackage{graphicx}
\usepackage{tensor}
\usepackage{textcomp}
\usepackage{xcolor}
\usepackage{subfigure} 
\usepackage{gensymb}
\usepackage{listings}
\usepackage{xcolor}
\usepackage{xspace}
\usepackage{float}
\def\BibTeX{{\rm B\kern-.05em{\sc i\kern-.025em b}\kern-.08em
    T\kern-.1667em\lower.7ex\hbox{E}\kern-.125emX}}
\begin{document}

\title{A Detection and Filtering Framework for Collaborative Localization\\
}
\author{\IEEEauthorblockN{\small{Thirumalaesh Ashokkumar}}
\IEEEauthorblockA{\textit{\small{University of Michigan}} \\
\small{Ann Arbor, Michigan}}
\and
\IEEEauthorblockN{\small{Katherine A. Skinner}}
\IEEEauthorblockA{\textit{\small{University of Michigan}} \\
\small{Ann Arbor, Michigan}}
\and
\IEEEauthorblockN{\small{Siddharth Agarwal}}
\IEEEauthorblockA{\textit{\small{IEEE Senior Member}}}
\and
\IEEEauthorblockN{\small{Ankit Vora}}
\IEEEauthorblockA{\textit{\small{IEEE Member}}}
\and
\IEEEauthorblockN{\small{Ashutosh Bhown}}
\IEEEauthorblockA{\textit{\small{University of Michigan}} \\
\small{Ann Arbor, Michigan}}

}
\maketitle

\begin{abstract}
\input{subsections/abstract}

\end{abstract}


\section{Introduction}
\input{subsections/intro.tex}
\section{Related Work}
\input{subsections/related_work.tex}
\section{Technical Approach}
\input{subsections/tech_app.tex}
\subsection{Localization}
\input{subsections/localization.tex}
\subsubsection{Extended Kalman Filter}
\input{subsections/ekf.tex} 
\subsubsection{Perception Simulator}
\input{subsections/Experimental_Setup.tex}

\subsubsection{Raw Poses}
\input{subsections/poses.tex}
\section{Experiments and Results}
\input{subsections/results.tex}
\section{Discussion}
\input{subsections/discussions.tex}
\section{Conclusion and Future Work}
\input{subsections/conclusion_and_future_work.tex}


\end{document}

%% file: subsections/abstract.tex
Increasingly, autonomous vehicles (AVs) are becoming a reality, such as the Advanced Driver Assistance Systems (ADAS) in vehicles that assist drivers in driving and parking functions with vehicles today. The localization problem for AVs relies primarily on multiple sensors, including cameras, LiDARs, and radars. Manufacturing, installing, calibrating, and maintaining these sensors can be very expensive, thereby increasing the overall cost of AVs. This research explores the means to improve localization on vehicles belonging to the ADAS category in a platooning context, where an ADAS vehicle follows a lead `Smart' AV equipped with a highly accurate sensor suite. We propose and produce results by using a filtering framework to combine pose information derived from vision and odometry to improve the localization of the ADAS vehicle that follows the smart vehicle. 



%% file: subsections/intro.tex
Autonomous Vehicles (AVs) are a reality waiting to happen, with broad applications in logistics, travel, and service industries. Moreover, the ability of AVs to collaborate, forming multi-agent networks would boost the productivity and efficiency of tasks performed by these vehicles. While the development of standalone autonomous vehicles has seen great strides, the development of multi-agent systems requires highly accurate mapping and localization of all agents. 

Due to cost and complexity, it is not always possible for all agents of a multi-agent autonomous vehicles network to possess equivalent sensing suites in terms of accuracy and precision. It would be ideal to exploit the sensing suites available on one vehicle with high-end sensors to improve the localization of other vehicles with a low-cost sensor suite. This premise is the focus of this work, where we will describe a mechanism of detecting and improving localization on a vehicle with a lower-grade sensor suite, by using the state estimations derived from a vehicle with a higher-grade sensor suite and relating these to the follower vehicle. 

\begin{figure}[t]
    \centering
    \includegraphics[width=\columnwidth, scale = 2]{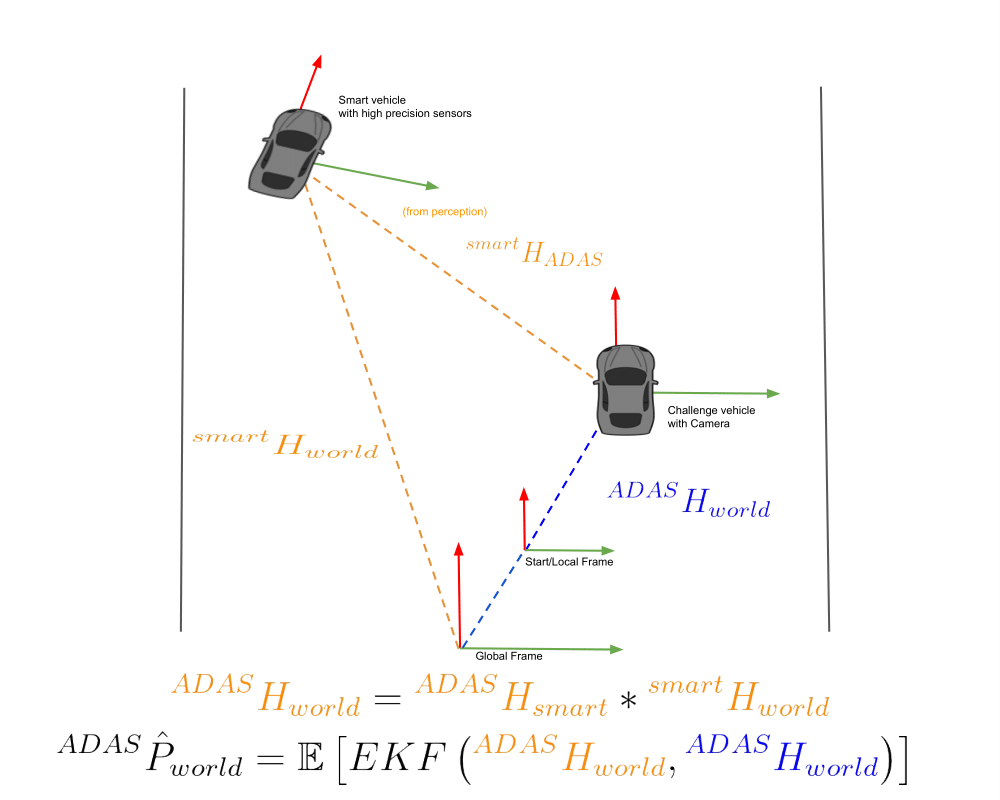}
    \caption{Improving localization of an 'ADAS' vehicle by fusing pose from inertial odometry({\color{blue} $^{ADAS}H_{world}$}) and pose information derived by perceiving the smart vehicle({\color{orange} $^{ADAS}H_{world}$}). Fusion is done through an EKF that gives the improved pose estimate $^{ADAS}\hat{P}_{world}$. $^AH_B$ is the pose of A in frame B.}
    \label{fig:Illustration}
\end{figure}

The specific focus of this work is a two-vehicle setup. The first is a lead vehicle having a robust and highly accurate sensor suite. We will refer to this vehicle as the smart vehicle. Improving the localization of the ADAS vehicle with diminished sensing capabilities is our primary goal. We propose a fusion framework to fuse the pose information acquired from the smart vehicle along with the odometry of the ADAS vehicle to improve its localization. We test our setup on the Ford Multi-AV Seasonal dataset \cite{b0}. The dataset contains odometry and other sensor data from  multiple vehicles driving through the Michigan-Detroit area. 
The illustration of our problem statement is shown in Fig \ref{fig:Illustration}.

The main contributions of this work are:
\begin{itemize}
    \item Development of a filtering framework for collaborative multi-agent localization
    \item Testing of the framework on a real-world dataset -  Ford Multi-AV Seasonal dataset
\end{itemize}

%% file: subsections/related_work.tex
While the solution to the localization problem on standalone vehicles can be viewed as part of the SLAM (Simultaneous Localization and Mapping) problem, the research into collaborative multi-agent localization takes varied approaches. A feature detection and two-stage filtering mechanism has been described in \cite{b4}. Another approach to solving this problem is to treat the multiple agents as parts of the same entity, which has been detailed by authors in \cite{b7}. An approach using particle filters, where the particles are shared between the agents has been proposed by \cite{b1}. \cite{b5} is a detailed survey that points out the various methods and problems related to Multi-Agent Localization. 

The literature on localization approaches can be classified into three broad categories: algorithms based on the Extended Kalman Filter, the Particle Filter, and Graph-based approaches. Solutions using a camera as the primary source of data are also being developed, such as solutions to the Visual SLAM problem. In this work, our goal is to establish the advantages of using a perception system on an ADAS vehicle with a lower-grade sensor suite to detect the smart vehicle with a higher-grade sensor suite, thereby improving the localization of the ADAS vehicle. Thus, for localization, we focus on using the Extended Kalman Filter and we make the assumption that there is a communications link between the two vehicles. 

%% file: subsections/tech_app.tex
\begin{figure}
    \centering
    \includegraphics[width=\columnwidth,scale=5]{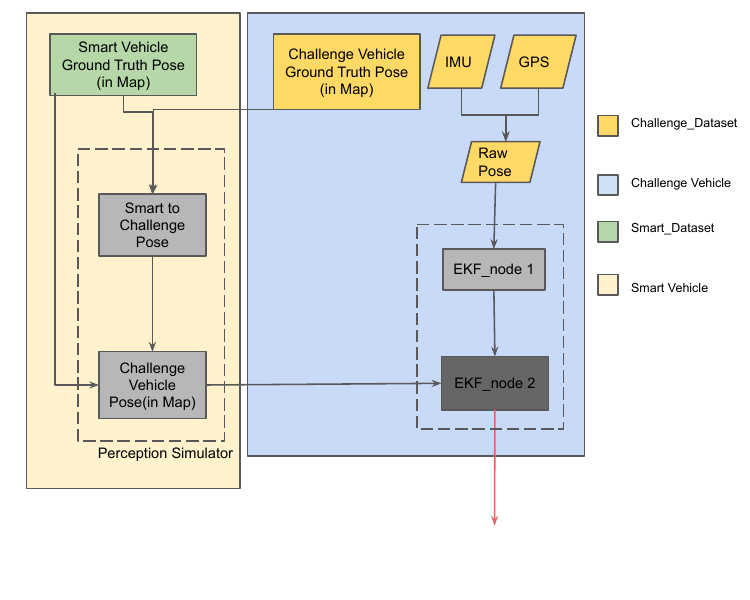}
    \vspace{-.5in}
    \caption{The proposed Framework: Perception module inputs a pose information to the filter, odometry information is also input to the network to obtain an improved pose estimate}
    \label{fig:fusion_framework}
\end{figure}

Consider the three important frames in Fig. \ref{fig:Illustration}. The Global frame$(H_{world})$ or Map frame is the frame from which all other frames are defined. This is usually dependent on the map being used. The map is considered to be known for the purposes of this work, and thus raw poses of both vehicles are defined in this frame. The Local or Start frame is the frame from which we begin the experiments. The measurements of the IMU are defined with respect to this frame. The Body frame is the frame of the vehicle in each instance.

For our experiments, since the map is as available from the dataset, we know the start position of the robot, implying we know the transformation between the Global frame and the Local frame, which is simply the start position of the vehicle. To perform filtering and get pose estimates in the Global frame, we need transforms from the Local frame to the Body frame at every instance of fusion, along with the transformation from the Global frame to the Local frame. Only then can the Global $\rightarrow{}$ Local $\rightarrow{}$ Body transform be achieved, through which the measurements can be fused.

To achieve the Local to Body transform at each instance, we use EKF\_node 1 as shown in Fig. \ref{fig:fusion_framework}, which fuses only the raw poses. Using the generated poses and a static transform to describe the Global to Local transform, we are able to use EKF\_node 2, to perform sensor fusion on the raw poses and the measurements from the perception module. Particularly, node 2 uses the poses of the ADAS vehicle($T_O^C$), and output of the perception module($T_W^C$) as input, fusing them based on the Extended Kalman Filter algorithm, to generate state estimates. We note that $T_O^C$ is fused relatively, while $T_W^C$ are fused as absolute measurements of the pose.

%% file: subsections/localization.tex
For localization and filtering, we use the filtering setup provided by the robot\_localization\cite{b12} ROS package. We set up two nodes of the Extended Kalman Filter for the reasons stated earlier. The first node, filters the pose odometry of the ADAS vehicle, to produce the transform between the start frame and the body frame. Using this, measurements of raw poses and perception are fused in the second node. The setup is described in detail below.

%% file: subsections/ekf.tex
The Extended Kalman filter (EKF) is a nonlinear estimation algorithm used to estimate the states of a system with nonlinear dynamics. It is an extension of the Kalman filter, which handles the nonlinearities by using a first-order linear approximation of the nonlinear system. The EKF works by linearizing the system around the current estimate and then updating the estimate using a combination of the linearized model and the measurements. We model the system as the linear approximation of a nonlinear system, given by\\
\begin{equation}
    X_{k+1} = A_kX_k + \xi, \xi \sim \mathcal{N}(0,\Sigma)
\end{equation}
where $A_k$ is the linear approximation of the posterior at step $k$. Here, $A_k$ used in the robot\_localization package is the standard motion model for a rigid body. We also note that by implication of using the EKF, we assume a zero mean Gaussian noise in the system dynamics, with covariance $\Sigma$ as shown above.

To fuse the odometry of the ADAS vehicle with the pose estimates from the smart vehicle, we consider a linear model for the measurements given by
\begin{equation}
    Y_k = H_kX_k + \eta, \eta \sim \mathcal{N}(0,\Gamma)
\end{equation}

Similar to the motion model, the noise here is considered to be zero mean Gaussian, with covariance given by $\Gamma$.


For the state estimation of the ADAS vehicle, we consider a 15 dimensional state vector, containing the positions, orientations, linear velocities, angular velocities and linear accelerations. The dataset contains IMU, GPS data and Pose data derived from the IMU and GPS data. Therefore, we would use the Pose data directly for fusion as the odometry data. Consequently, the measurement model is simplified to being an identity transformation. Based on the frequency of the odometry measurements, pose estimates from the smart vehicle, we carry out the prediction and update steps sequentially. 

%% file: subsections/Experimental_Setup.tex
We simulate the perception module by using the noisy ground truth data. Using the ground truth pose of the ADAS vehicle ($P_W^C$) and ground truth pose of the smart vehicle ($P_W^S$), we arrive at the pose of the ADAS vehicle in the frame of the smart vehicle ($P_S^C$). 
\begin{equation}
    P_S^C = P^C_WP^W_S
\end{equation}
During the simulations, independent noises are added to the translation along $x$ and $y$ ($T$) and rotation ($R$) of these poses, such that:
\begin{equation}
  T' = T + \Sigma, \Sigma \sim(0, \sigma)
\end{equation}
We work with rotations as quaternions, and rotation noise is added such that,
\begin{equation}
    R' = \begin{bmatrix} x \\ y \\ z \\ w \end{bmatrix} * \begin{bmatrix}
        0 \\ 0 \\ cos(\theta/2) \\ sin(\theta/2) \end{bmatrix} , \theta \sim( 0 , \gamma)
\end{equation}
where, $\theta$ is the rotation about z. This transformation is then combined back with the pose of the smart vehicle in the world frame ($P_W^S$) to give us The new pose of the ADAS vehicle in the world frame ($T_W^C$). The translation noises are added to only $x$,$y$ coordinates.
\begin{equation}
  P_S^{''C} = \left [\begin{array}{c|c}
  R' & T'\end{array} \right]
\end{equation}
\begin{equation}
    T^C_W = P^{''C}_S P^S_W 
\end{equation}
This measurement $T_C^W$ is used as the input to the second EKF node.

%% file: subsections/poses.tex
Raw poses are taken from the rosbag, which contains position and orientation information along with timestamps. These are fused pose estimates from the Applanix POS module, which uses its internal IMU, GPS and estimator. Given that the accuracy of the sensors are high, we will perturb these measurements with noises, supplanting for a noisy sensor. For translations $T_{raw}$ and rotation quaternions $R_{raw}$ from the bag,
\begin{equation}
    T_{raw}^{'} = T_{raw} + \Sigma_{raw} , \Sigma \sim(0, \sigma_{raw})
\end{equation}
\begin{equation}
    R'_{raw} = \begin{bmatrix} x \\ y \\ z \\ w \end{bmatrix} * \begin{bmatrix}
        0 \\ 0 \\ cos(\theta_{raw}/2) \\ sin(\theta_{raw}/2) \end{bmatrix} , \theta_{raw} \sim( 0 , \gamma_{raw})
\end{equation}
\\

%% file: subsections/results.tex
\subsection{Dataset} The Ford Multi-AV Seasonal dataset contains the sensor data in rosbags. These rosbags contain data in the NED (North-East-Down) frame, and this is converted to the ENU (East-North-Up) frame, to align with the standards set by ROS REP-103 for outdoor navigation. Vehicle 1 is considered the smart vehicle, while vehicle 2 is the ADAS vehicle. Given the internal clocks of the vehicles have not been synchronized, we perform this synchronization using GPS time. We compute and publish $T_C^W$ only if the time difference between when $P_W^S$ and $ P_W^C$ is received is less than a threshold of 0.1s. For purposes of this work, we perform filtration on a specific portion of the dataset, that consists of two vehicles in proximity to each other. The results shown below are derived using the data from Vehicle 1 and Vehicle 2 from the logs of $24^{th}$ July 2017. 
\subsection{Results} The experiments are structured to test the filter performance through various noise levels and frequencies of data availability. During the experiments, it was noted that the pose estimates along the $z$-axis for ADAS vehicle (Vehicle 2), were very noisy and uncharacteristic of the sensor specifications. 

\begin{figure}[H]
    \includegraphics[width=1\columnwidth,scale=1]{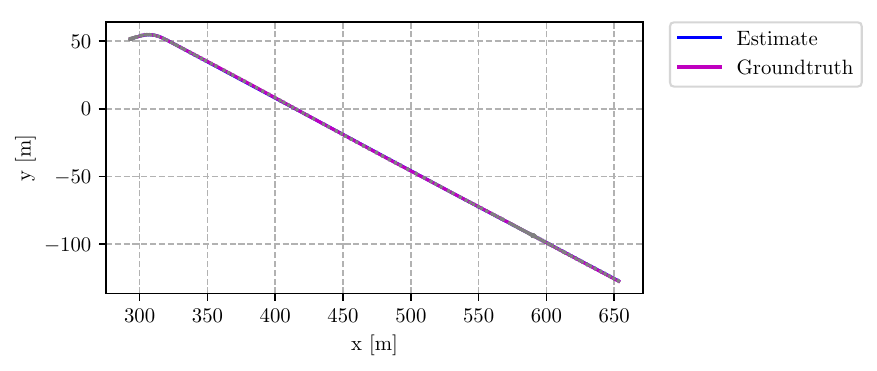}
    \caption{Trajectory of ADAS Vehicle, $\sigma$=5, $\gamma$=5, no added noise on raw pose}
    \label{fig:Trajectory of ADAS Vehicle_basic}
\end{figure}

The frequency of the ADAS vehicle and smart vehicle's poses is around 200 Hz. In the first experiment, we simulate the perception module to also produce estimates using all of the measurements. However, we do add noises to the perception module measurements as described in previous sections. We set $\sigma = 5, \gamma = 5$, to simulate the noises that would arise from using visual feedback. Using this and the raw pose (odometry) measurements of the ADAS vehicle, we produce 6DoF state estimates for the ADAS vehicle, whose trajectory is shown in Fig \ref{fig:Trajectory of ADAS Vehicle_basic}. Note that the raw poses are only corrupted by the noise inherent in the sensor, Applanix POS and its internal estimator. We use the rpg\_trajectory evaluation tool \cite{b13} on ROS to perform trajectory alignment and analysis. 
\begin{figure}[H]
    \includegraphics[width=1\columnwidth,scale=1]{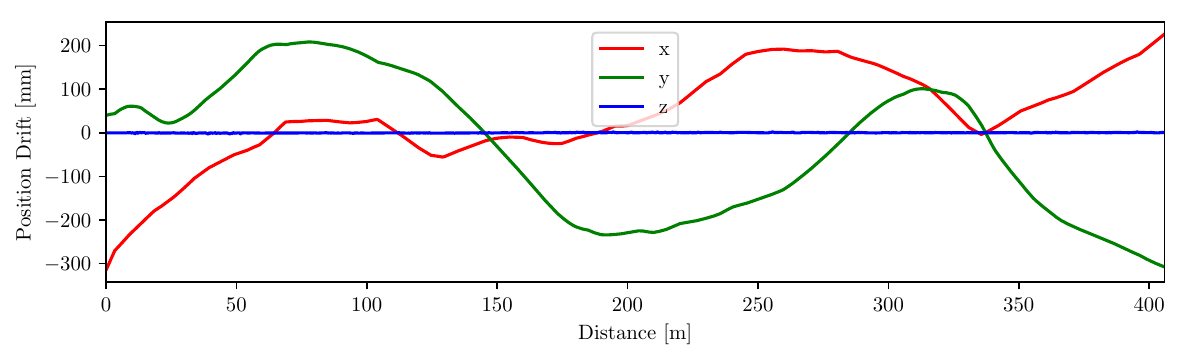}
    \caption{Position error ADAS Vehicle, $\sigma$=5, $\gamma$=5, no added noise on raw pose}
    \label{fig:Trajectory of ADAS Vehicle pos_basic}
\end{figure}\begin{figure} 
    \includegraphics[width=1\columnwidth,scale=1]{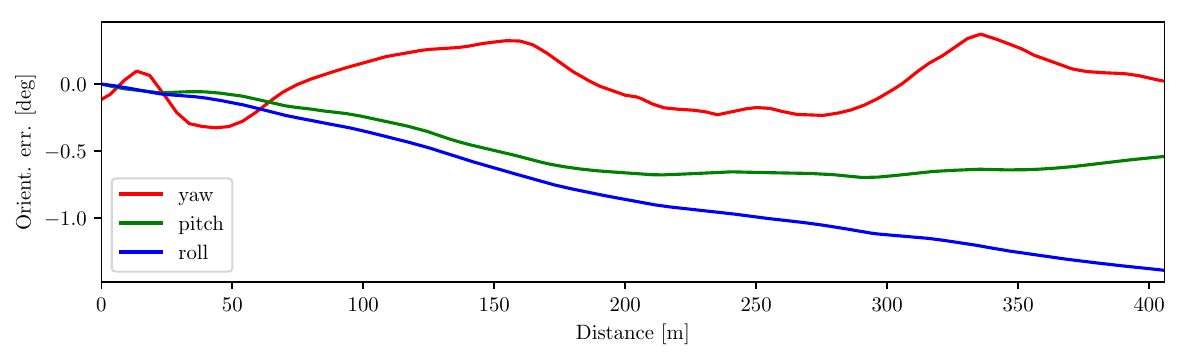}
    \caption{Orientation error of ADAS Vehicle,, $\sigma$=5, $\gamma$=5, no added noise on raw pose}
    \label{fig:Trajectory of ADAS Vehicle orie_basic}
\end{figure}
In Fig. \ref{fig:Trajectory of ADAS Vehicle_basic}, the estimated pose vs. ground truth pose is presented. The corresponding error is shown in Fig. \ref{fig:Trajectory of ADAS Vehicle pos_basic} and Fig. \ref{fig:Trajectory of ADAS Vehicle orie_basic}. As expected, it is evident that when the odometry is accurate, the noises from the perception measurements are filtered out, and we get an absolute mean translation accuracy of 0.127m and an absolute mean orientation accuracy of 1.080 degrees.

In the next experiment, we add noises to the raw pose measurements. We perform ablation on various noise levels across each state. For purposes of this analysis, we study the effects of various noise levels in the perception module, for a particular noise in the raw pose measurements. Tables \ref{2.5_2.5_0} and \ref{2.5_2.5_1} show the errors in translation at different perception noise levels and the error with only the odometry. Note that fixing $ \sigma_{raw} = 2.5m$, $ \gamma_{raw} = 0\degree$ corresponds to having an accurate heading with a high translation noise, where 68\% of the additive noises are within (-2.5, 2.5).



\begin{table}[H]
\caption{Translation RMSE for $ \sigma_{raw} = 2.5m$, $ \gamma_{raw} = 0\degree$}
\centering
\begin{tabular}{|c|c|c|c|c|c|c|}
  \hline
   Noises Std Dev &  {$\sigma = 0.3m $} & {$\sigma = 0.6m $} & {$\sigma = 0.9m $} \\ 
   \hline
   w/o perception & \multicolumn{3}{c|}{0.846} \\ \hline
   $\gamma = 10\degree$ &  0.095 & 0.155 & 0.183 \\ 
   \hline
   $\gamma = 15\degree$ &  0.097 & 0.144 & 0.185 \\  
   \hline
\end{tabular}
\label{2.5_2.5_0}
\end{table}

\begin{table}[H]
\caption{Translation RMSE for $ \sigma_{raw} = 2.5m$, $ \gamma_{raw} = 1\degree$}
\centering
\begin{tabular}{|c|c|c|c|c|c|c|}
  \hline
   Noises Std Dev &  {$\sigma = 0.3m $} & {$\sigma = 0.6m $} & {$\sigma = 0.9m $} \\ 
   \hline
    w/o perception & \multicolumn{3}{c|}{1.311} \\ \hline
   $\gamma = 10\degree$ &  0.111 & 0.230 & 0.217 \\ 
   \hline
   $\gamma = 15\degree$ &  0.123 & 0.235 & 0.250 \\  
   \hline
\end{tabular}
\label{2.5_2.5_1}
\end{table}
However, in practice, the perception module is not expected to produce data at such a high frequency. While the above experiments showcase the potential of using the perception module, to be a practical solution, we need to test its performance at lower frequencies. To sub-sample the data in the rosbag, we make use of the drop tool in ROS. For the experiment, the frequency of the raw messages is dropped down to approximately 100 Hz, while the frequency of the smart vehicle's pose information is dropped down to approximately to 5 Hz. And the results of this are shown in Table \ref{2.5_2.5_0.5}.

\begin{table}[H]
\caption{Translation RMSE for $ \sigma_{raw} = 2.5m$, $ \gamma_{raw} = 0.5\degree$}
\centering
\begin{tabular}{|c|c|c|c|c|c|c|}
  \hline
   Noises Std Dev &  {$\sigma = 0.3m $} & {$\sigma = 0.6m $} & {$\sigma = 0.9m $} \\ 
   \hline
    w/o perception & \multicolumn{3}{c|}{1.061} \\ \hline
   $\gamma = 10\degree$ &  0.318 & 0.825 & 0.560 \\ 
   \hline
   $\gamma = 15\degree$ &  0.446 & 0.641 & 0.499 \\  
   \hline
\end{tabular}
\label{2.5_2.5_0.5}
\end{table}

\begin{figure}[H]
    \centering
    \includegraphics[width=1\columnwidth,scale=1]{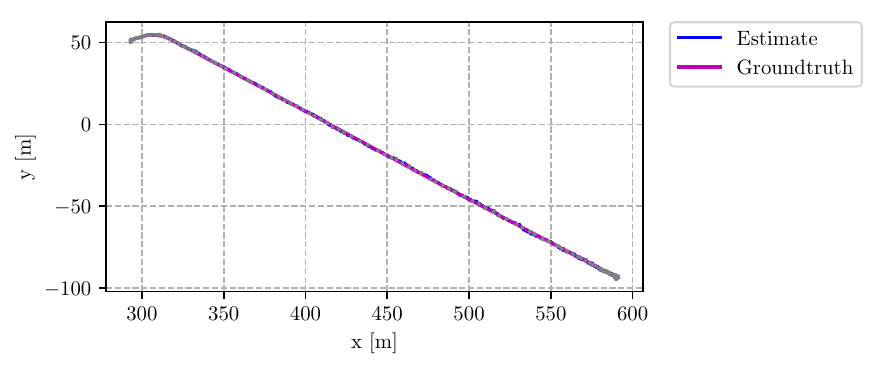}
    \caption{Trajectory for $\gamma = 10\degree, \sigma = 0.3m$, reduced frequency}
    \label{fig:0.5traj}
\end{figure}

As an example, Fig. \ref{fig:0.5traj} shows the trajectory of the vehicle along with the ground truth at one of the noise levels, where we can see a near smooth trajectory.


%% file: subsections/discussions.tex
From the experiments, it is evident that, even in the presence of noisy odometry, the use of the perception module can improve the overall localization of the ADAS vehicle. Furthermore, the improvement in translation is more prevalent for a wider range of raw pose noises. Using perception does improve heading estimates in some cases. However, it fails to do so when the odometry heading noise levels are beyond a bound. This is likely because given poor heading estimates from both sensor inputs, the filter is unable to filter out the heading noises, which worsens when integrated over time. We find that results are promising even when the frequencies of the measurements are low, as shown in Table \ref{2.5_2.5_0.5}.
Another behavior that was noticed was high heading errors with high oscillations in the beginning, which settled to a lower value after a period of time. 

%% file: subsections/conclusion_and_future_work.tex
In this work, we have developed and presented a localization mechanism that can be used for multi-agent localization, using odometry and visual feedback. The experiments show a clear and consistent improvement in translational accuracy, and bounded errors. The framework when tested on the Ford Multi-AV Seasonal Dataset is robust over various noise levels. 

With a working localization pipeline, the future scope of this work is to use a detection and association system instead of the simulator. This would give us a complete detection and filtering framework that can be deployed and tested in real-world systems. Timestamp matching for the data was done using a constant offset which was calculated with GPS time as the reference. However, a better approach can be used to further synchronize the data from both vehicles accurately. Also, in terms of the filter, we have not deployed any outlier rejection techniques or applied any constraints on the evolution of the system dynamics, so this is left to future work.

%% file: conference_101719.bbl
\begin{thebibliography}{00}
\bibitem{b0} Agarwal S, Vora A, Pandey G, Williams W, Kourous H, McBride J. Ford Multi-AV Seasonal Dataset. The International Journal of Robotics Research. 2020;39(12):1367-1376. doi:10.1177/0278364920961451
\bibitem{b4} Z. Zhou, W. Tang, Z. Wang, L. Wang and R. Zhang, "Multi-robot Real-time Cooperative Localization Based on High-speed Feature Detection and Two-stage Filtering," 2021 IEEE International Conference on Real-time Computing and Robotics (RCAR), Xining, China, 2021, pp. 690-696, doi: 10.1109/RCAR52367.2021.9517423.
\bibitem{b7} Roumeliotis, S.I., Bekey, G.A. (2000). Distributed Multi-Robot Localization. In: Parker, L.E., Bekey, G., Barhen, J. (eds) Distributed Autonomous Robotic Systems 4. Springer, Tokyo. https://doi.org/10.1007/978-4-431-67919-6\_17
\bibitem{b1} Xian-lun TANG, La-mei LI, Bo-jie JIANG, Mobile robot SLAM method based on multi-agent particle swarm optimized particle filter, The Journal of China Universities of Posts and Telecommunications, Volume 21, Issue 6, 2014, Pages 78-86.
\bibitem{b5} Saeedi S, Trentini M, Seto M, Li H. Multiple-robot simultaneous localization and mapping: A review. Journal of Field Robotics, 2016, 33(1): 3–46 DOI:10.1002/rob.21620
\bibitem{b12} Moore T, Stouch D, A Generalized Extended Kalman Filter Implementation for the Robot Operating System, Proceedings of the 13th International Conference on Intelligent Autonomous Systems (IAS-13), July 2014
\bibitem{b13} Zhang, Zichao and Scaramuzza, Davide, A Tutorial on Quantitative Trajectory Evaluation for Visual(-Inertial) Odometry, IEEE/RSJ Int. Conf. Intell. Robot. Syst. (IROS), 2018
\bibitem{b2} P. Biber and W. Strasser, "The normal distributions transform: a new approach to laser scan matching," Proceedings 2003 IEEE/RSJ International Conference on Intelligent Robots and Systems (IROS 2003) (Cat. No.03CH37453), Las Vegas, NV, USA, 2003, pp. 2743-2748 vol.3, doi: 10.1109/IROS.2003.1249285.
\bibitem{b3} Danping Zou, Ping Tan, Wenxian Yu, Collaborative visual SLAM for multiple agents:A brief survey, Virtual Reality \& Intelligent Hardware, Volume 1, Issue 5, 2019, Pages 461-482
\bibitem{b6} Fox, D., Burgard, W., Kruppa, H., Thrun, S. (1999). Collaborative Multi-Robot Localization. In: Förstner, W., Buhmann, J.M., Faber, A., Faber, P. (eds) Mustererkennung 1999. Informatik aktuell. Springer, Berlin, Heidelberg. https://doi.org/10.1007/978-3-642-60243-6\_2
\bibitem{b8} Nadia Nedjah, Luiza Macedo Mourelle, Pedro Jorge Albuquerque de Oliveira, Simultaneous localization and mapping using swarm intelligence based methods, Expert Systems with Applications, Volume 159, 2020, 113547.
\bibitem{b9}Takashi Matsubara, Masao Kubo \& Yusuke Murachi (2010) Particle Filter for Collaborative Multi-Robot Localization Tolerant to Recognition Error, Advanced Robotics, 24:15,2043-2058, DOI: 10.1163/016918610X534259
\bibitem{b16} Reid, Tyler \& Houts, Sarah \& Cammarata, Robert \& Mills, Graham \& Agarwal, Siddharth \& Vora, Ankit \& Pandey, Gaurav. (2019). Localization Requirements for Autonomous Vehicles. 
\end{thebibliography}
